%% file: main.tex
\documentclass[runningheads]{llncs}
\usepackage{graphicx}
\usepackage{hyperref}
\newcommand{\JimSan}{Jim\'{e}nez-S\'{a}nchez}

\usepackage{subcaption}
\usepackage{amsmath}
\usepackage{tabularx}
\usepackage{graphicx}
\usepackage{adjustbox}
\usepackage[T1]{fontenc}

\usepackage{color}
\definecolor{myred}{rgb}{.8,.0,.0}

\usepackage{forest}

\begin{document}
\title{Source Matters: Source Dataset Impact on Model Robustness in Medical Imaging}
\titlerunning{Source Matters}
%
\author{Dovile Juodelyte\inst{1} \and Yucheng Lu\inst{1} \and Amelia \JimSan\inst{1} \and Sabrina Bottazzi\inst{2} \and Enzo Ferrante\inst{3} \and Veronika Cheplygina\inst{1}}

\authorrunning{D. Juodelyte et al.}
%
\institute{IT University of Copenhagen, Denmark\\
\email{\{doju,yucl,amji,vech\}@itu.dk} \and
Universidad Nacional de San Martín, Argentina\\
\email{sbottazzi@estudiantes.unsam.edu.ar} \and
CONICET - Universidad Nacional del Litoral, Argentina\\
\email{eferrante@sinc.unl.edu.ar } }


%
\maketitle              
\begin{abstract}
Transfer learning has become an essential part of medical imaging classification algorithms, often leveraging ImageNet weights. The domain shift from natural to medical images has prompted alternatives such as RadImageNet, often showing comparable classification performance. However, it remains unclear whether the performance gains from transfer learning stem from improved generalization or shortcut learning. To address this, we conceptualize confounders by introducing the Medical Imaging Contextualized Confounder Taxonomy (MICCAT) and investigate a range of confounders across it -- whether synthetic or sampled from the data -- using two public chest X-ray and CT datasets. We show that ImageNet and RadImageNet achieve comparable classification performance, yet ImageNet is much more prone to overfitting to confounders. We recommend that researchers using ImageNet-pretrained models reexamine their model robustness by conducting similar experiments. Our code and experiments are available at \url{https://github.com/DovileDo/source-matters}.

\end{abstract}

\section{Introduction} \label{sec:intro}
\input{sec1_introduction}

\section{Method} \label{sec:method}

\input{sec3_method_taxonomy_dovile}
\input{sec3_method_confounders}

\section{Results and Discussion} \label{sec:results}
\input{sec4_results.tex}

\section{Conclusion} \label{sec:discussion}
\input{sec5_discussion}

\begin{credits}

\subsubsection{\ackname} This study was funded by the Novo Nordisk Foundation (grant number NNF21OC0068816). We acknowledge the NIH Clinical Center (NIH CXR14 data), National Cancer Institute and the Foundation for the National Institutes of Health and their critical role in the creation of the free publicly available LIDC/IDRI Database used in this study.


\subsubsection{\discintname} The authors have no competing interests to declare that are relevant to the content of this article.
\end{credits}

%
%

\bibliographystyle{splncs04}
\bibliography{refs_manual}


\end{document}

%% file: sec1_introduction.tex
Machine learning models hold immense promise for revolutionizing healthcare. However, their deployment in real-world clinical settings is hindered by various challenges, with one of the most critical being their hidden reliance on spurious features \cite{wiens2019no}. Recent research has highlighted the detrimental effects of this reliance, including bias against demographic subgroups \cite{banerjee2023shortcuts}, limited generalization across hospitals \cite{zech2018variable}, and the risk of clinical errors that may harm patients \cite{oakden2020hidden}.

Despite transfer learning becoming a cornerstone in medical imaging, its impact on model generalization remains largely unexplored. Pre-training on ImageNet has become a standard practice due to its success in 2D image classification. While some studies have explored alternative medical source datasets for pre-training  \cite{cheplygina2019cats,mei2022radimagenet,zhou2021models,juodelyte2023revisiting}, ImageNet continues to serve as a strong baseline.

Recent literature suggests that the size of the source dataset may matter more than its domain or composition  \cite{ramanujan2024connection,gavrikov2022does}. However, \cite{jain2023databased} demonstrated performance improvements through source dataset pruning. In this context, we argue that cross-domain transfer can be problematic, especially when source dataset selection is solely based on classification performance, as it may inadvertently lead to shortcut learning rather than genuine improvements in generalization. Shortcut learning can be considered antithetical to generalization and robustness as it is not a failure to generalize per se, but rather a failure to generalize in the intended direction \cite{geirhos2020shortcut}.

In this paper, we investigate how the domain of the source dataset affects model generalization. First, we conceptualize confounding factors in medical images by introducing the Medical Imaging Contextualized Confounder Taxonomy (MICCAT) and generate synthetic or sample real-world confounders from MICCAT, commonly found in chest X-rays and CT scans, to systematically assess model robustness. Second, we compare models pre-trained on natural (ImageNet) and medical (RadImageNet) datasets across X-ray and CT tasks and show substantial differences in robustness to shortcut learning despite comparable predictive performance. While transfer learning has been observed to enhance model robustness \cite{hendrycks2019using}, our results suggest that it may not hold true when transferring across domains, cautioning against using ImageNet pre-trained models in medical contexts due to their susceptibility to shortcut learning. Furthermore, our findings highlight the limitations of conventional performance metrics based on i.i.d. datasets, which fail to discern between genuine improvements in generalization and shortcut learning. Thus, we advocate for a more nuanced evaluation of transfer learning effectiveness to ensure the reliability and safety of machine learning applications in clinical settings.

%% file: sec3_method_taxonomy_dovile.tex
\begin{figure}[t!]
    \centering
    \includegraphics[width=\textwidth]{{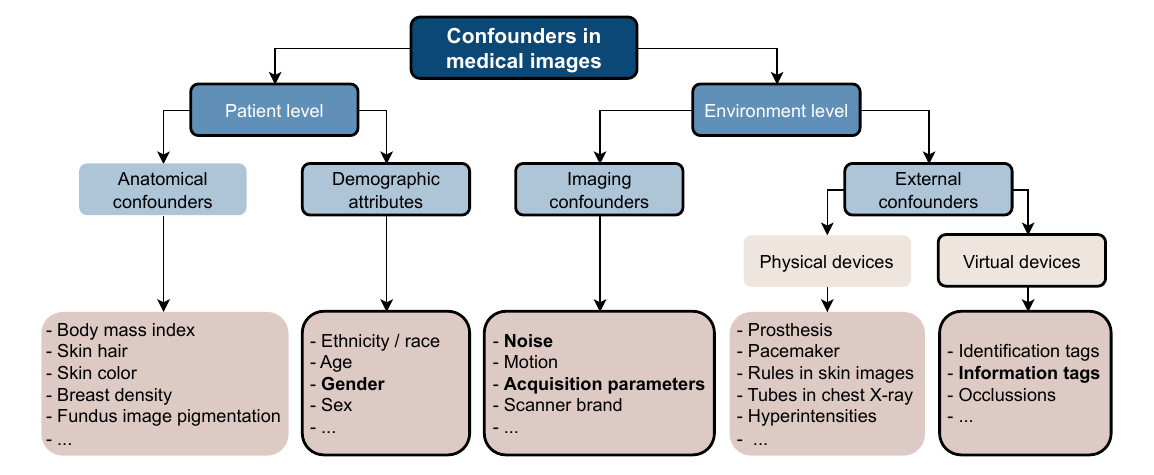}}
    \caption{\textbf{MICCAT}: Medical Imaging Contextualized Confounder Taxonomy. Instances of confounders investigated in this paper are highlighted in bold.}
    \label{fig:miccat}
\end{figure}

\subsection{MICCAT: towards a standardized taxonomy for medical imaging confounders}
To the best of our knowledge, there is no standardized taxonomy for classifying potential confounders in medical images. Thus, to better structure our robustness analysis, we propose a new taxonomy: Medical Imaging Contextualized Confounder Taxonomy (MICCAT).

Previous work has shown that standard demographic attributes such as sex, age, or ethnicity may act as confounders, leading to shortcut learning and potentially disadvantaging historically underserved subgroups \cite{banerjee2023shortcuts}. However, solely focusing on standard protected demographic attributes may overlook other specific factors related to clusters of patients for which the systems tend to fail \cite{voneuler-chelpin2019sensitivity}. In MICCAT, we identify these as `contextualized confounders', as they are often domain or context-specific, associated with particular image modalities, organs, hospitalization conditions, or diseases.

First, MICCAT differentiates between \textit{patient level} and \textit{environment level} confounders. At the \textit{patient level}, we make a distinction between standard \textit{demographic attributes} (e.g., sex, age, race) and contextualized \textit{anatomical confounders}, which arise from inherent anatomical properties of the organs and human body or disease variations in images. This distinction is crucial as standard demographic attributes often serve as proxies for underlying causes of learned shortcuts. For instance, ethnicity may proxy skin color in dermatoscopic images. Identifying the true shortcut cause allows for more targeted interventions to mitigate biases. We define the concept of \textit{environment level} confounders, which stem from contextualized \textit{external} or \textit{imaging confounders}. The former include physical or virtual elements in images due to external factors like hospitalization devices or image tags, while the latter include characteristics related to the imaging modality itself, such as noise, motion blur, or differences in intensities due to equipment or acquisition parameters. Fig. \ref{fig:miccat} illustrates this taxonomy with examples for each category.

\noindent \textbf{Confounders studied in this paper.} We explore the MICCAT by investigating four examples of confounders, highlighted by a black outline in Fig. \ref{fig:miccat}:
\begin{itemize}
\item An external confounder (\emph{a tag}) placed in the upper left corner of the image, representing confounding features introduced by various imaging devices across or within hospitals (Fig. \ref{fig:R}).
\item Two typical imaging confounders: \emph{denoising} (Fig. \ref{fig:low}), widely used by various vendors to reduce noise for enhanced readability \cite{hasegawa2022noise}, and \emph{Poisson noise} (Fig. \ref{fig:noise}), originating from quantum statistics of photons, which cannot be mitigated through hardware engineering, unlike noise introduced by circuit-related artifacts \cite{wei2021physics}.
\item A patient-level confounder where we use \emph{patient gender}, which is easily accessible in metadata, as a proxy for a broader spectrum of anatomical confounders. We use the same term for this variable as in the original dataset.
\end{itemize} 

%% file: sec3_method_confounders.tex
\subsection{Experimental Design}\label{sec:design}

We investigate the impact of source dataset domain on model generalization by comparing ImageNet \cite{deng2009imagenet} and RadImageNet \cite{mei2022radimagenet} models, which are fine-tuned using binary prediction tasks for findings in open-access chest X-ray (NIH CXR14 \cite{nih}) and CT (LIDC-IDRI \cite{IDRI}) datasets curated to include systematically controlled confounders. NIH CXR14 is used to represent cross-domain transfer for both ImageNet and RadImageNet, as X-ray is not included in RadImageNet, while LIDC-IDRI serves as an in-domain example for RadImageNet and a cross-domain example for ImageNet.

\begin{figure}[t]
    \centering
    \begin{subfigure}[t]{0.298\textwidth}
        \centering
        \includegraphics[width=\textwidth]{{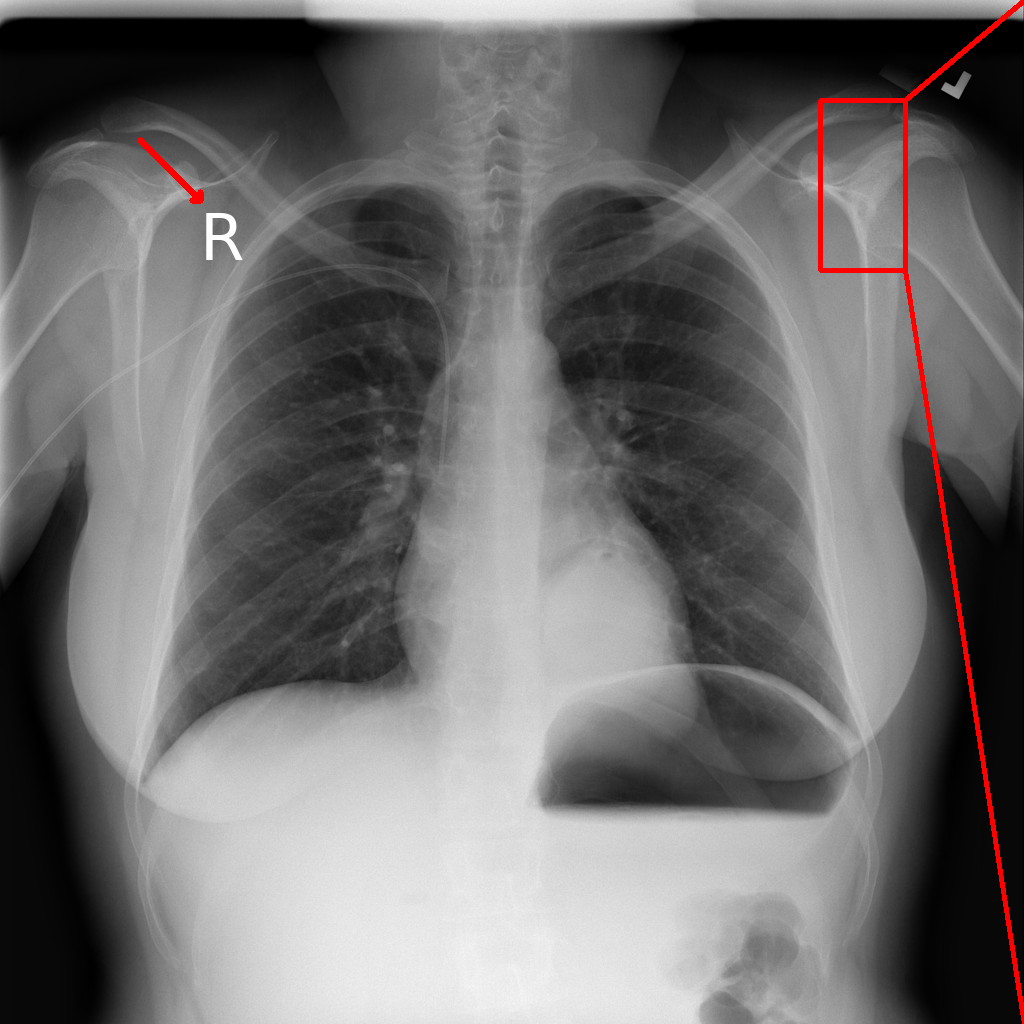}}
        \caption{}
        \label{fig:R}
    \end{subfigure}%
    \begin{subfigure}[t]{0.149\textwidth}
        \centering
        \includegraphics[width=\textwidth]{{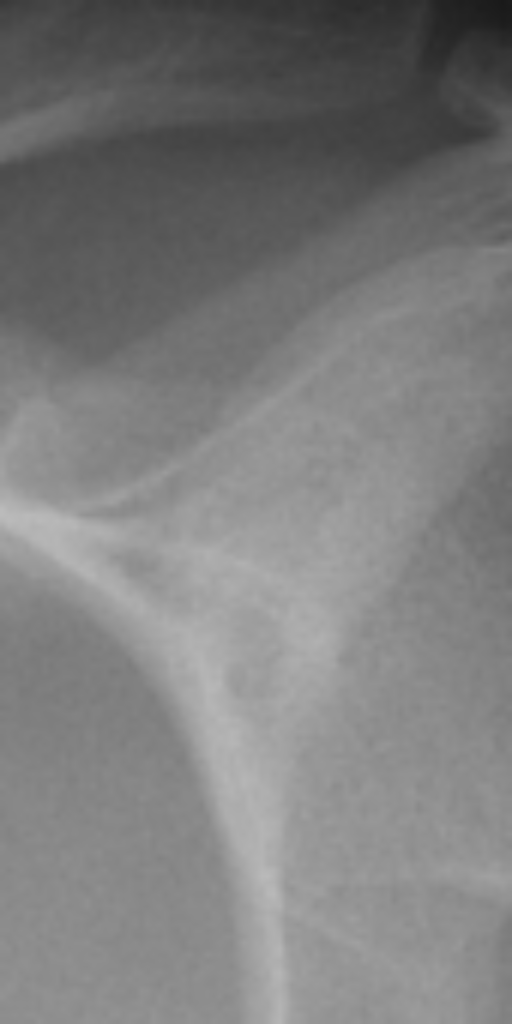}}
        \caption{}
    \end{subfigure}%
    \begin{subfigure}[t]{0.149\textwidth}
        \centering
        \includegraphics[width=\textwidth]{{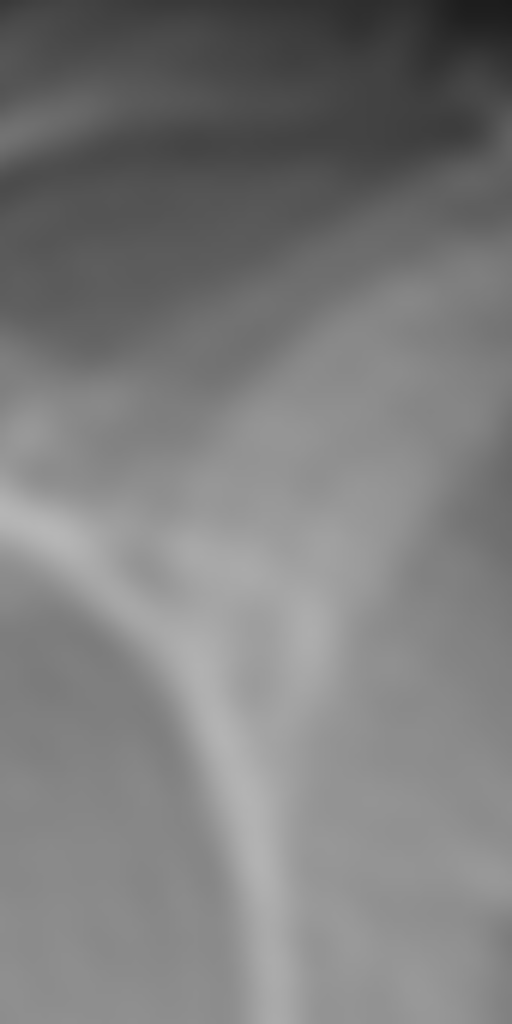}}
        \caption{}
        \label{fig:low}
    \end{subfigure}%
    \begin{subfigure}[t]{0.149\textwidth}
        \centering
        \includegraphics[width=\textwidth]{{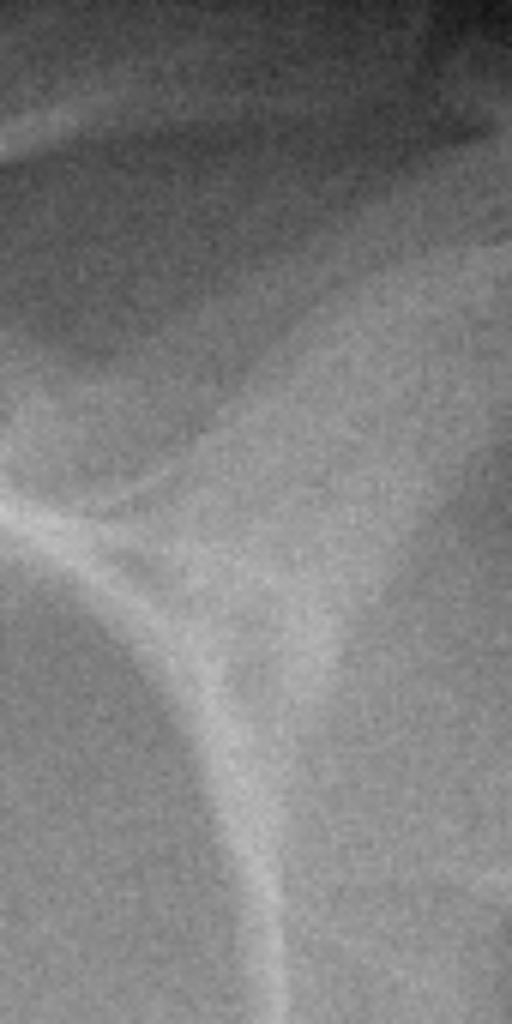}}
        \caption{}
        \label{fig:noise}
    \end{subfigure}
    \caption{\textbf{Synthetic artifacts}: (a) \textit{A tag} with a red arrow for reference, (b) a zoomed-in view of the original image, (c) \textit{Denoising} by low-pass filter with cutoff frequency (see Eq. \ref{eq:low}) of $D_0 = 200$px, and (d) \textit{Poisson noise} with $N_{0}= 2 \times 10^{6}$ (see Eq. \ref{eq:noise}). The parameters used here are to emphasize subtle local variations such as the smoothing effect of the low-pass filter and the graininess introduced by the Poisson noise. For our experiments, we use $D_0 = 500$px and $N_{0}= 2\times 10^{7}$ which are imperceptible.}
\end{figure}

\noindent \textbf{Confounder generation.} 
\textit{Patient gender} is sampled to correlate `Female' with the label.

\textit{A tag} is placed further away from the edges (starting at $200\times200$px in the original image of $1024\times1024$px), to ensure it remains intact during training despite augmentations applied (Fig. \ref{fig:R}).

The simplest method for \textit{Denoising} is applying low-pass filtering which entails converting the input image from the spatial to the frequency domain using Discrete Fourier Transform (DFT), followed by element-wise multiplication with the low-pass filter $H_{LPF}(u,v)$ to generate the filtered image:

\begin{equation}
  H_{LPF}(u,v) =\left\{
    \begin{array}{ll}
        1, & D(u,v)\leq D_0\\
        0, & \mbox{otherwise}
    \end{array}
  \right.
  \label{eq:low}
\end{equation}

where $D(u,v)$ represents the distance from the origin in the frequency domain, and $D_0$ is the specified cutoff frequency. In our experiments, we set $D_0 = 500$px. Subsequently, the high-frequency suppressed image is reconstructed in the spatial domain via the Inverse Discrete Fourier Transform (IDFT), resulting in a smoothing effect (see Fig. \ref{fig:low}).

\textit{Poisson noise} originating from quantum statistics of photons is formulated as a Poisson random process:
\begin{equation}
    \left ( p_{r} + N_{p} \right ) =  \mathcal{P}\left ( p_{r} \right )
    \label{eq:noise}
\end{equation}
where $N_{p}$ represents Poisson noise, which notably affects image quality under low-dose conditions (e.g., low-dose CT and X-ray screenings), while the linear recording $p_{r} = \exp\left (-p_{a}  \right )N_{0}$ is obtained via the reversed conversion from attenuation $p_{a}$ given the prior information of the source intensity $N_{0}$, where $p_{a}$ is the pixel values of projections, obtained from the image space as described in \cite{leuschner2021lodopab}.
To simulate low-dose screening, we add Poisson noise to the image (Fig. \ref{fig:noise}) by adjusting the $N_{0}$ parameter to control noise levels. We aim for minimal noise, setting $N_{0}= 2 \times 10^{7}$ after visually examining the noise to ensure it remains imperceptible.

\noindent \textbf{Evaluation.} 
To investigate shortcut learning systematically, we construct development datasets for fine-tuning, focusing on a binary classification task. We introduce previously mentioned confounders (e.g., `Female') into the positive class with a controlled probability $p_{\text{art}} \in \{0, 0.1, 0.2, 0.5, 0.8, 1\}$ to deliberately influence the learning process, replicating scenarios where real-world data may contain confounders. To assess the presence of shortcut learning, we evaluate the fine-tuned models with independently and identically distributed (i.i.d.) as well as out-of-distribution (o.o.d.) test sets. In the o.o.d. set, we introduce the same artifact used during fine-tuning to the negative class with $p_{\text{art}} = 1$, such that the models are tested on instances where artifacts appear in the opposite class compared to what they encountered during training. We evaluate the fine-tuned models using the AUC (area under the receiver operating characteristic curve).

\setlength{\tabcolsep}{5pt}
\begin{table}[t]
\centering
\caption{Target datasets used for fine-tuning. T: \textit{tag}, D: \textit{denoising}, N: \textit{noise}.}
\label{tab:data}
\begin{adjustbox}{width=\textwidth}
\begin{tabular}{@{}lccccrr@{}}
\hline
&
 &
  \multicolumn{1}{c}{\textbf{\# images in}} &
  \multicolumn{1}{c}{\textbf{\% split}} &
  \multicolumn{1}{c}{\textbf{\% class split}} &
  \multicolumn{1}{c}{\textbf{Image}} &
  \multicolumn{1}{c}{\textbf{Batch}}\\ 
\textbf{Task} &
\textbf{Confounder} &
  \multicolumn{1}{c}{\textbf{test/dev(train$+$val)}} &
  \multicolumn{1}{c}{\textbf{train/val}} &
  \multicolumn{1}{c}{\textbf{pos/neg}} &
  \multicolumn{1}{c}{\textbf{size}} &
  \multicolumn{1}{c}{\textbf{size}}  \\ \hline
         Lung mass (NIH CXR14~\cite{nih}) & T, D, N & 83/248 & 90/10 &  30/70 & 512 $\times$ 512 & 32\\ 
        Lung mass (LIDC-IDRI~\cite{IDRI}) & T, D, N & 1710/500 & 80/20 & 50/50 & 362 $\times$ 362 & 32 \\ 
         Atelectasis (NIH CXR14~\cite{nih}) & Gender & 400/400 & 85/15 & 50/50 & 256 $\times$ 256 & 64 \\
    \hline
\end{tabular}
 \end{adjustbox}

\end{table}

\noindent \textbf{Medical targets.} We create separate binary classification tasks for lung mass detection using subsets of images sourced from two datasets: the chest X-ray NIH CXR14 \cite{nih} subset annotated by clinicians \cite{nabulsi2021deep}, and the chest CT dataset LIDC-IDRI \cite{IDRI} annotated by four radiologists. From the latter, we sample paired positive and negative 2D slices from the original 3D scans using nodule ROI annotations, representing any kind of lesions and their nearby slices without remarkable findings. We include synthetic artifacts (\textit{a tag, denoising,} and \textit{Poisson noise}) in both tasks. For the case where patient gender serves as the confounding feature, we sample posterior to anterior (PA) images from NIH CXR14 to construct a binary classification task for atelectasis. We deliberately limit the size of our development datasets, encompassing both balanced and unbalanced class distributions to cover a spectrum of clinical scenarios. Data splits for training, validation, and testing preserve class distribution and are stratified by patient. Further details are available in Table \ref{tab:data}.

\noindent \textbf{Fine-tuning details.}
We use ResNet50 \cite{he2016deep}, InceptionV3 \cite{szegedy2016rethinking}, InceptionResNetV2 \cite{szegedy2017inception}, and DenseNet121 \cite{huang2017densely} as the backbones with average pooling and a dropout layer (0.5 probability). The models are trained using cross-entropy loss with Adam optimizer (learning rate: $1 \times 10^{-5}$) for a maximum of 200 epochs with early stopping after 30 epochs of no improvement in validation loss (AUC for the balanced tasks). This configuration, established during early tuning, proved flexible enough to accommodate different initializations and target datasets. During training, we apply image augmentations including random rotation (up to 10 degrees), width and height shifts, shear, and zoom, all set to 0.1, with a fill mode set to `nearest'. Models were implemented using Keras \cite{chollet2015keras} library and fine-tuned on an NVIDIA Tesla A100 GPU card.

%% file: sec4_results.tex
\begin{figure}[t!]
    \centering
    \includegraphics[width=0.9\textwidth]{{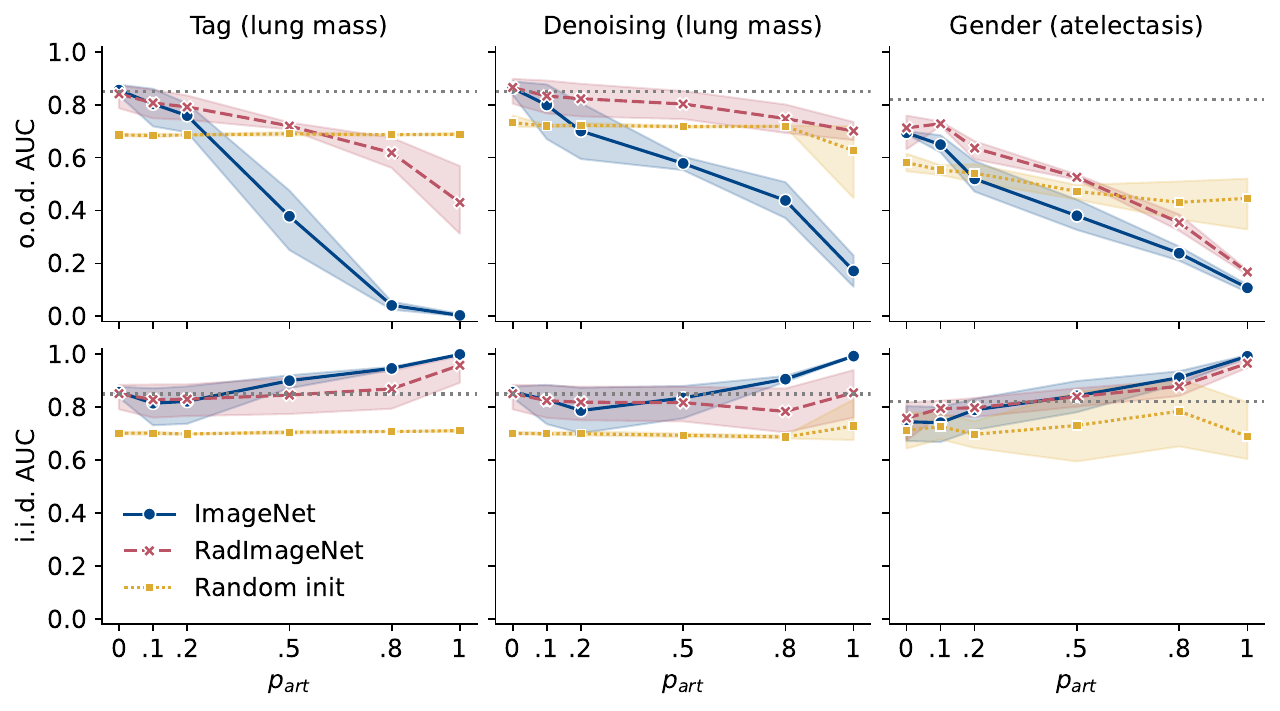}}
    \caption{Mean AUC across five-fold cross-validation with 95\% CI for lung mass (left and middle) and atelectasis (right) prediction in chest X-rays. Increasing correlation between artifact (\textit{tag}, \textit{denoising}, \textit{gender}) and the label leads to lower o.o.d. AUC (on o.o.d. test set as described in Sec. \ref{sec:design}) (top row), while i.i.d. AUC increases (bottom row). RadImageNet pretraining shows less degradation in o.o.d. AUC compared to ImageNet pretraining, suggesting that ImageNet may over-rely on spurious correlations in the target dataset. The grey dotted line is the SOTA result for lung mass and atelectasis in NIH CXR14 reported by \cite{dai2024unichest}.}
    \label{fig:results}
\end{figure}

\textbf{RadImageNet is robust to shortcut learning.} Fig. \ref{fig:results} shows that ImageNet and RadImageNet achieve comparable AUC on i.i.d. test set, however, when subjected to o.o.d. test set, notable differences emerge. Specifically, ImageNet's o.o.d. performance on X-rays, confounded by \textit{tag, denoising,} and \textit{patient gender}, drops more compared to RadImageNet, indicating ImageNet's higher reliance on spurious correlations. This could be because certain features, for instance, \textit{a tag} (letters), may serve as a discriminative feature in ImageNet, e.g., for the computer keyboard class. However, RadImageNet is invariant to such features as they are not consistently associated with specific labels across different classes, and this invariance transfers to the target task. We observed similar trends in the CT dataset, with the o.o.d. AUC decreasing from 0.84 to 0.02 for ImageNet, and to 0.22 for RadImageNet (for \textit{tag}); and from 0.7 to 0.01 for ImageNet, and from 0.83 only to 0.6 for RadImageNet (for \textit{denoising}). It is worth noting that RadImageNet models tend to train longer, averaging 141 epochs across all experiments, compared to 72 epochs for ImageNet models.

\begin{figure}[t]
    \centering
    \includegraphics[width=0.65\textwidth]{{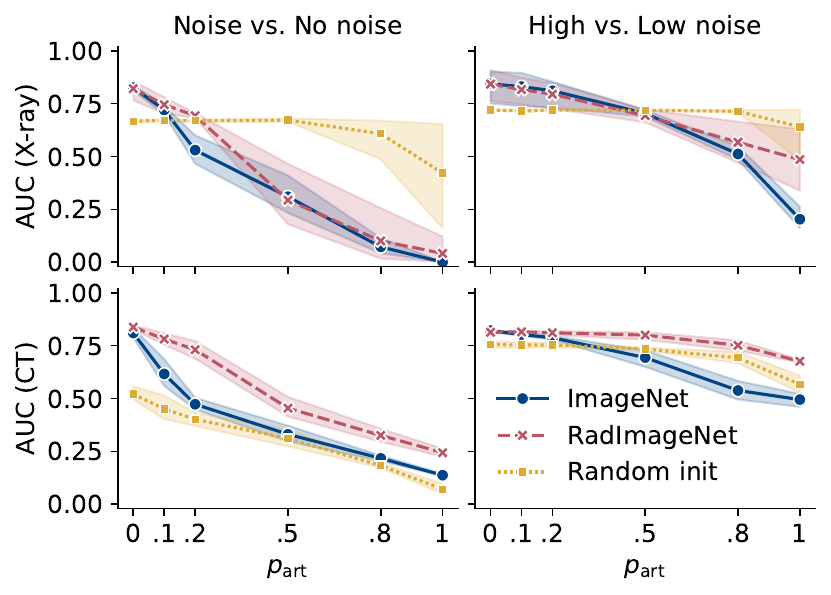}}
    \caption{O.o.d. AUC (mean and 95\% CI across five-folds) for lung mass prediction in chest X-rays and CTs. In X-rays (top), both ImageNet and RadImageNet show similar reliance on Poisson noise. However, RadImageNet is more robust in CT scans (bottom). When the confounder is high vs low noise, both ImageNet and RadImageNet are less sensitive (right), compared to noise vs no noise (left).}
    \label{fig:results_noise}
\end{figure}

Although \textit{tag} and \textit{denoising} are designed to replicate real-world artifacts, they lack the diversity found in real-world scenarios. \textit{Patient gender} presents a more realistic confounder. Here, the performance gap between ImageNet and RadImageNet is smaller (by 0.12 on average for $p_{\text{art}} \geq 0.1$) yet remains statistically significant (permutation test, $0.008<p\text{-value}<0.032$, for $p_{\text{art}}\geq 0.1$). 
This suggests that RadImageNet's resilience to shortcuts extends to more realistic confounder variations, further emphasizing its robustness in medical image classification. Here we only provide results for ResNet50,
however, we observed similar results for InceptionV3, InceptionRes-NetV2, and DenseNet121.

Random initialization appears robust to shortcut learning, with consistent o.o.d. performance as $p_{\text{art}}$ increases. However, this is mainly due to the unbalanced class distribution in the lung mass prediction task within the NIH CXR14 dataset, where randomly initialized models tend to predict the overrepresented negative class ($\text{recall}=0$). Conversely, in the case of a balanced class distribution in the CT target dataset, the o.o.d. performance of randomly initialized models deteriorates to a similar degree as that of ImageNet-initialized models.

\noindent \textbf{Shortcuts come in all shapes and sizes.} ImageNet and RadImageNet both heavily rely on Poisson noise in X-rays (Fig. \ref{fig:results_noise}, upper left) but RadImageNet shows greater robustness to noise in CT scans compared to ImageNet (Fig. \ref{fig:results_noise}, lower left). It is important to note that Poisson noise manifests differently in X-rays and CT scans. In X-rays, Poisson noise introduces graininess characterized by random and pixel-wise independent variations, while in CT scans, it appears as streak artifacts structurally correlated to projections and thus is not pixel-wise independent in the image domain. 

To understand the impact of this difference, we directly introduce Poisson noise $N_{0}= 2 \times 10^{7}$ in the image domain for CT scans, mimicking the pixel-wise independence seen in X-rays. However, since CT scans inherently contain noise, this introduces a confounding feature of high versus low levels of noise, as opposed to the original confounder of noise versus no noise.

To simulate a corresponding scenario in X-rays, we generate two levels of Poisson noise: $N_{0}= 2 \times 10^{7}$ for the positives and $N_{0}= 1 \times 10^{7}$ for the negatives (reversed for the o.o.d. test set). Both models show a smaller drop in o.o.d. AUC across modalities, indicating a reduced reliance on the noise shortcut (Fig. \ref{fig:results_noise}, right). This suggests that discerning between high and low noise levels is a more challenging task than simply detecting the presence of noise. 

RadImageNet maintains its robustness in CT scans, while in X-rays, RadImageNet relies on noise to a similar extent as ImageNet. This may be explained by the absence of X-ray images in RadImageNet, leading to a lack of robust X-ray representations that would resist pixel-wise independent noise -- a phenomenon less common in CT, MR, and ultrasound, modalities included in RadImageNet. This highlights that even transferring from a medical source of a different modality may lead to overfitting on confounders.

While our findings generalize over the four tested CNNs, we did not investigate other architectures, such as transformers, due to CNNs competitive performance \cite{doerrich2024rethinking}. Although we expect that our observations might hold true for transformers, given their tendency to reuse features to an even greater extent than CNNs \cite{matsoukas2022makes}, we defer experimental verification to future research.

In our exploration of the MICCAT, we found that RadImageNet models are generally more robust to shortcuts. However, there is some variability within the category of \textit{imaging confounders}, and the importance of the source domain in \textit{anatomical confounders} seems to be lower. Expanding the scope to include other confounders would offer a more comprehensive understanding of the taxonomy landscape and provide insights into the nuances within each category, facilitating better-informed source dataset selection and evaluation strategies. MICCAT paves the way for a more systematic approach to addressing shortcut learning in medical imaging in general by providing a framework for thorough confounder curation and enabling a comprehensive analysis.

%% file: sec5_discussion.tex
Our study sheds light on the critical role of the source dataset domain in generalization in medical imaging tasks. By systematically investigating confounders typically found in X-rays and CT scans, we uncovered substantial differences in robustness to shortcuts between models pre-trained on natural and medical image datasets. Our findings caution against the blind application of transfer learning across domains. We advocate for a more nuanced evaluation to improve the reliability and safety of machine learning applications in clinical settings.

\noindent \textbf{Prospect of application.} Transfer learning plays a fundamental role in machine learning applications for medical imaging. Our study emphasizes the often underestimated importance of selecting pre-trained models, urging a necessary reevaluation and deeper investigation into their use in clinical practice.
